\pdfoutput=1

\documentclass[11pt]{article}

\usepackage[]{acl}

\usepackage{times}
\usepackage{latexsym}

\usepackage[T1]{fontenc}
\usepackage{amsfonts}

\usepackage[utf8]{inputenc}

\usepackage{microtype}

\usepackage{amsmath}
\usepackage{amssymb}
\usepackage{graphicx}

\usepackage{float}
\usepackage{pgfplots}
\usepackage{subfigure}
\usepackage{booktabs}
\usepackage{multicol}
\usepackage{multirow}
\usepackage{arydshln} 
\usepackage{CJKutf8}

\title{Solution of DeBERTaV3 on CommonsenseQA}

\date{}

\author{Letian Peng$^{1,2,3}$, Zuchao Li$^{1,2,3}$, and Hai Zhao$^{1,2,3}$\thanks{$\ $  Corresponding author.}\\
$^{1}$Department of Computer Science and Engineering, Shanghai Jiao Tong University \\
	$^{2}$Key Laboratory of Shanghai Education Commission for Intelligent Interaction \\ and Cognitive Engineering, Shanghai Jiao Tong University, Shanghai, China\\
	$^{3}$MoE Key Lab of Artificial Intelligence, AI Institute, Shanghai Jiao Tong University \\
  {\tt \small \{zxc-00,charlee\}@sjtu.edu.cn, zhaohai@cs.sjtu.edu.cn}}

\begin{document}
\maketitle
\begin{abstract}

We report the performance of DeBERTaV3 on CommonsenseQA in this report. We simply formalize the answer selection as a text classification for DeBERTaV3. The strong natural language inference ability of DeBERTaV3 helps its single and ensemble model set the new (w/o external knowledge) state-of-the-art on CommonsenseQA. 

\end{abstract}

\section{Task and Dataset}

CommmonsenseQA\footnote{\href{https://www.tau-nlp.org/commonsenseqa}{https://www.tau-nlp.org/commonsenseqa}} \cite{DBLP:conf/naacl/TalmorHLB19} is a commonsense question answering dataset that requires the model to select an answer from five candidates. The negative choices are sampled based on the graphs in ConceptNet \cite{DBLP:conf/acl/LiTTG16}. 

CommonsenseQA contains $9741$, $1221$, $1140$ questions (totally $12102$) in its train, dev, and test datasets. 

\section{Methodology}

DeBERTa \cite{he2021deberta} is a pre-trained language model with an enhanced decoding procedure. DeBERTaV3 \cite{he2021debertav3} refines the training process by replacing the initial training objective, masked language modeling, with replaced token detection. 

We formalize the question selection as a text classification by transforming a question-answer pair into the following prompt.

\begin{center}
    \textit{Q [SEP] A.}
\end{center}

\noindent where, \textit{Q}, \textit{A}, \textit{[SEP]} refer to the question, answer and separation token. For instance,

\begin{center}
    \textit{Where can you find all of space? [SEP] Universe.}
\end{center}

We use a text classifier with DebertaV3 as the backbone to score the prompt. We score the five answer candidates of a question and use cross-entropy loss with the correct label as the objective to train the model. 

\section{Configuration\footnote{Our code is here:\\ \href{https://github.com/Stareru/CSQA_DeBERTaV3}{https://github.com/Stareru/CSQA\_DeBERTaV3}}}

We set the batch size to $8$, the initial learning rate to $10^{-5}$ with a linear decay of $0.67$ for each $5000$ steps. The training is run for $4$ epochs, and we report the best results on the dev dataset\footnote{We send our predictions on the test dataset to the leaderboard during writing this report.}. We create an ensemble model using the average scores of $5$ models on each answer candidate. 

\section{Experiment Results}

\begin{table}
\centering
\scalebox{1.0}{
\begin{tabular}{lcc}
\toprule
Method & Single & Ensemble\\
\midrule
ALBERT+MSKF$^\dag$ & 84.4 & - \\
ALBERT+DESC-KCR$^\dag$ & 84.7 & - \\
\midrule
RoBERTa & 78.5 & - \\
RoBERTa+FreeLB & 78.8 & - \\
ALBERT & 81.2 & 83.7 \\
ALBERT+HeadHunter & 83.3 & - \\
\midrule
DeBERTa$_{\textrm{base}}$ & 60.3 & 62.2 \\
DeBERTa$_{\textrm{large}}$ & 76.5 & 78.8 \\
DeBERTaV3$_{\textrm{base}}$ & 78.7 & 79.6 \\
DeBERTaV3$_{\textrm{large}}$ & \textbf{84.1} & \textbf{85.3}  \\
\bottomrule
\end{tabular}
}
\caption{Experiments Results on CommonsenseQA.\\ $\dag$: The method uses external knowledge.} 
\label{tab:res}
\end{table}

We report the experiments with DeBERTaV3$_{\textrm{base}}$ and DeBERTaV3$_{\textrm{large}}$. For comparison, we retrieve results from baselines \citep{DBLP:conf/acl/XuZXLZH21,DBLP:conf/iclr/ZhuCGSGL20,DBLP:conf/emnlp/LiZLAHZ21} on the top of the leaderboard that reported their results on the development dataset. ALBERT+DESC-KCR and ALBERT+MSKF are two baselines that respectively use ConceptNet and Wikidictionary as external knowledge base. To show the benefits from the replaced token detection objective, we add the initial DeBERTa as another baseline. 

Table~\ref{tab:res} shows DeBERTaV3 outperforms all baselines without external knowledge in the leaderboard and reaches the new state-of-the-art. With an exhaustively simple training scenario, DeBERTa can perform better than models with complex architecture. Also, DeBERTaV3 reaches a close performance to models that use external knowledge base, showing its strong capability in natural language inference. Compared to the initial DeBERTa, Replaced token detection leads to a sharp improvement in CommonsenseQA, especially on the base model.

\section{Conclusion}

This report presents the performance on CommonsenseQA of DeBERTaV3, which sets the new (w/o external knowledge) state-of-the-art. We also verify the benefits of replacing the masked language model with the replaced token detection training objective in DeBERTaV3. 

\bibliography{acl_latex}
\bibliographystyle{acl_natbib}

\end{document}